\pdfoutput=1
\documentclass{article}

\usepackage[final]{neurips_2024}

\usepackage[utf8]{inputenc} 
\usepackage[T1]{fontenc}    
\usepackage{hyperref}       
\usepackage{url}            

\usepackage{booktabs}       
\usepackage{amsfonts}       

\usepackage{nicefrac}       
\usepackage{microtype}      
\usepackage{xcolor}         
\usepackage{physics} 
\usepackage{dsfont} 
\usepackage[capitalise]{cleveref} 
\usepackage[pdftex]{graphicx}
\usepackage{algorithm}
\usepackage{algpseudocode}
\usepackage{algorithmicx}

\title{Sparse patches adversarial attacks via extrapolating point-wise information}

\author{%
  Yaniv Nemcovsky 
  \thanks{Department of Computer Science; Technion -- Israel Institute of Technology}\\
  \texttt{yanemcovsky@campus.technion.ac.il} \\
  \AND
  Avi Mendelson $^*$\\
  \texttt{mendlson@technion.ac.il} \\
  \And
  Chaim Baskin
  \thanks{School of Electrical and Computer Engineering; Ben-Gurion University of the Negev}\\
  \texttt{chaimbaskin@bgu.ac.il} \\
}
\begin{document}

\maketitle
\begin{abstract}
Sparse and patch adversarial attacks were previously shown to be applicable in realistic settings and are considered a security risk to autonomous systems. Sparse adversarial perturbations constitute a setting in which the adversarial perturbations are limited to affecting a relatively small number of points in the input. Patch adversarial attacks denote the setting where the sparse attacks are limited to a given structure, i.e., sparse patches with a given shape and number. However, previous patch adversarial attacks do not simultaneously optimize multiple patches' locations and perturbations. This work suggests a novel approach for sparse patches adversarial attacks via point-wise trimming of dense adversarial perturbations. Our approach enables simultaneous optimization of multiple sparse patches' locations and perturbations for any given number and shape. Moreover, our approach is also applicable for standard sparse adversarial attacks, where we show that it significantly improves the state-of-the-art over multiple extensive settings. A reference implementation of the proposed method and the reported experiments is provided at \url{https://github.com/yanemcovsky/SparsePatches.git}.
\end{abstract}    
\section{Introduction}
\label{sec:intro}

Adversarial perturbations were first discovered in the context of deep neural networks (DNNs), where the networks' gradients were used to produce small bounded-norm perturbations of the input that significantly altered their output \cite{szegedy2013intriguing}. Methods for optimizing such perturbations and the resulting perturbed inputs are denoted as adversarial attacks and adversarial inputs. Such attacks target the increase of the model's loss or the decrease of its accuracy and were shown to undermine the impressive performance of DNNs in multiple fields.
The norm bounds on adversarial perturbations are usually discussed in either the $L_\infty$ or $L_2$ norms \cite{szegedy2013intriguing,goodfellow2014explaining,madry2018towards}. Sparse adversarial attacks, in contrast, are a setting 
where $L_0$ norm bounds are applied and limit the perturbations to affect a relatively small number of points in the input. Sparsity $L_0$ norm bounds can also be applied in addition to the usually considered norms of $L_\infty,L_2$ but we consider such out of the scope of the current work. \cite{croce2019sparse,fan2020sparse,croce2021mind,dong2020greedyfool}. Patch adversarial attacks are a sub-setting of sparse attacks, where the perturbed points are constrained to constitute patches of a given shape and number. Patch adversarial attacks are highly realistic and were shown to be applicable in multiple real-world settings \cite{nemcovsky2022physical,xu2019evading,zolfi2021translucent,wei2022simultaneously,chen2019shapeshifter}. However, the optimization of sparse adversarial patches is computationally complex and entails the simultaneous optimization of the patches' locations and corresponding perturbations. Moreover, the locations' optimization is not directly differentiable and mandates a search over combinatorial spaces that grow exponentially with the number of patches. Previous patch attacks do not solve this optimization but rather a problem relaxation. Such attacks either optimize the perturbations over fixed locations \cite{nemcovsky2022physical,chen2019shapeshifter}, optimize the locations of fixed patches \cite{wei2022adversarial,zolfi2021translucent}, or limit the optimization to be over a single patch \cite{wei2022simultaneously}.
In contrast, previous sparse attacks that do not discuss patches suggest several approaches for simultaneously optimizing the selection of points to perturb and point-wise perturbations. To solve this complex optimization problem, \citet{modas2019sparsefool} first suggested approximating the non-convex $L_0$ norm by the convex $L_1$ norm, proposing the $SparseFool (SF)$ attack. Following this, \citet{croce2019sparse} suggested to utilize binary optimization and presented the $PGD_{L0}$ PGD-based \cite{madry2018towards} attack. \citet{goodfellow2020generative} then suggested first increasing the number of perturbed points, then reducing any unnecessary, presenting the $GreedyFool (GF)$ attack. Lastly, \citet{zhu2021sparse} suggested a homotopy algorithm and the $Homotopy$ attack.

In the present work, we suggest a novel approach for simultaneously optimizing multiple sparse patches' locations and perturbations. Our approach is based on point-wise trimming of dense adversarial perturbations and enables the optimization of patches for any given number and shape. To the best of our knowledge, this is the first direct solution to the complex optimization problem of adversarial patches. Moreover, our solution does not require differentiability during the trimming process and is therefore applicable to all the real-world settings presented in previous works \cite{nemcovsky2022physical,xu2019evading,zolfi2021translucent,wei2022simultaneously,chen2019shapeshifter}. In all these settings, our solution enables the optimization to be over a more extensive scope of patch adversarial attacks. In addition, our approach applies to standard sparse adversarial attacks, and we compare it to previous works on the $ImageNet$ classification task over various models. We consider $\epsilon_0$ bounds up to the common sparse representation bound of root input size \cite{elad2010sparse} and show that we significantly outperform the state-of-the-art for all the considered settings. 

\section{Background}
\label{sec:related}
\begin{figure*}
    \centering
    \resizebox{\linewidth}{!}{
    \begin{tabular}{ccccc}  
    \textbf{Clean}&\textbf{Dense perturbation} & \textbf{Trim to $\epsilon_0=32768$} & \textbf{Trim to $\epsilon_0=8192$} & \textbf{Final trim to $\epsilon_0=224$}\\
    \includegraphics[width=0.2\textwidth]{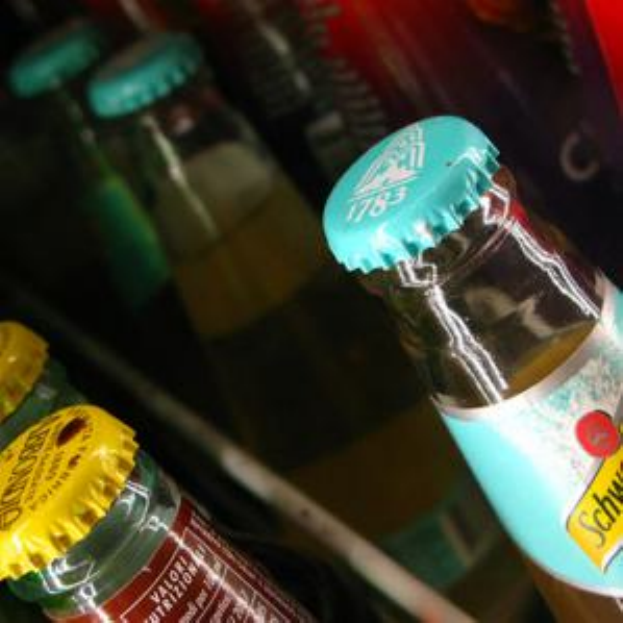}$$&
    \includegraphics[width=0.2\textwidth]{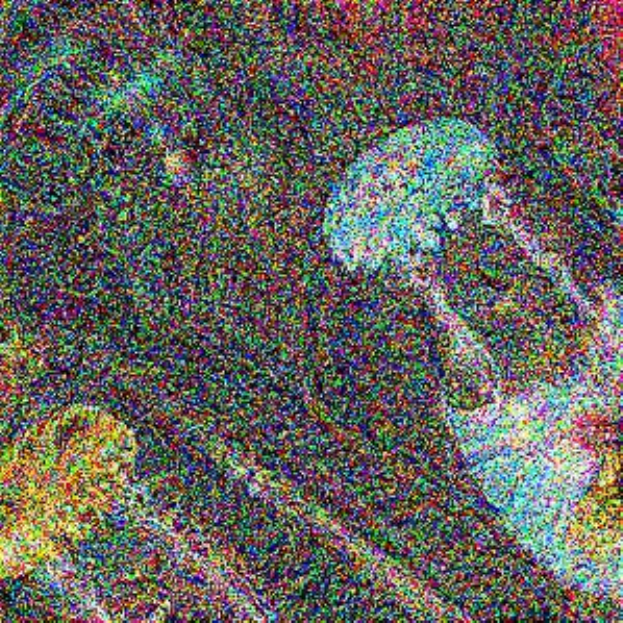}$$&
    \includegraphics[width=0.2\textwidth]{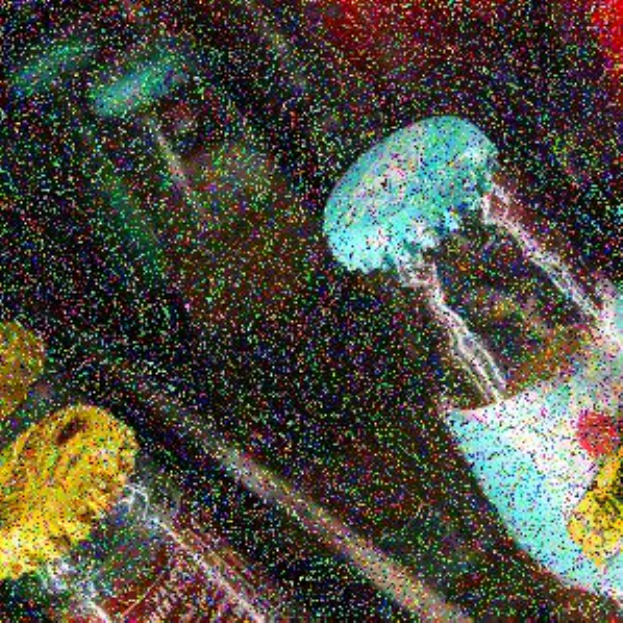}$$&
    \includegraphics[width=0.2\textwidth]{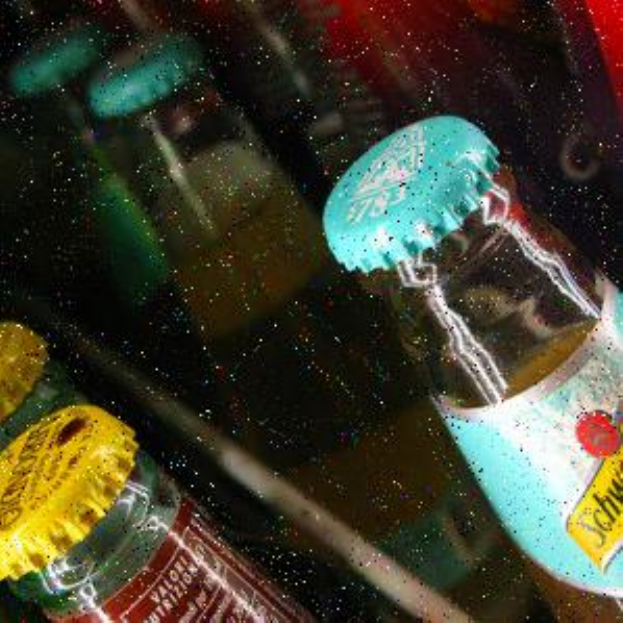}$$&
    \includegraphics[width=0.2\textwidth]{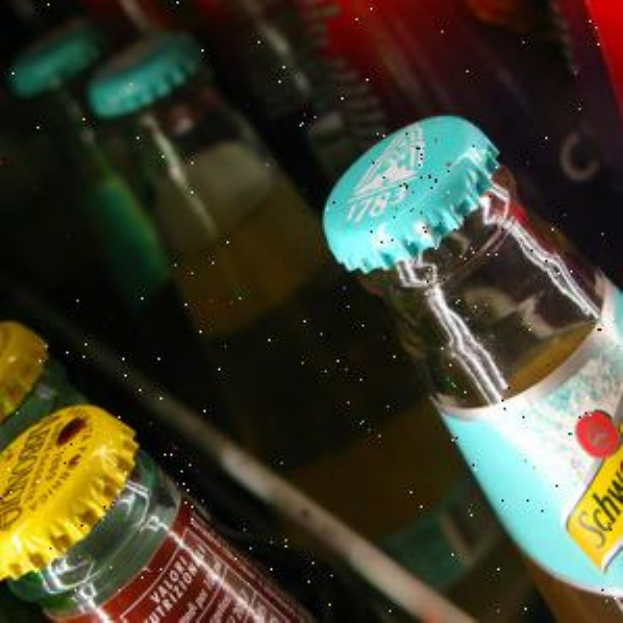}$$
    \\
    True label: \textbf{pop bottle}&
    Predicted label: \textbf{jellyfish}&
    Predicted label: \textbf{jellyfish}&
    Predicted label: \textbf{jellyfish}&
    Predicted label: \textbf{jellyfish} \\
    \includegraphics[width=0.2\textwidth]{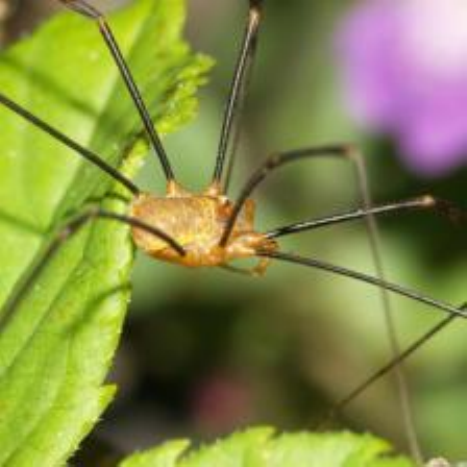}$$&
    \includegraphics[width=0.2\textwidth]{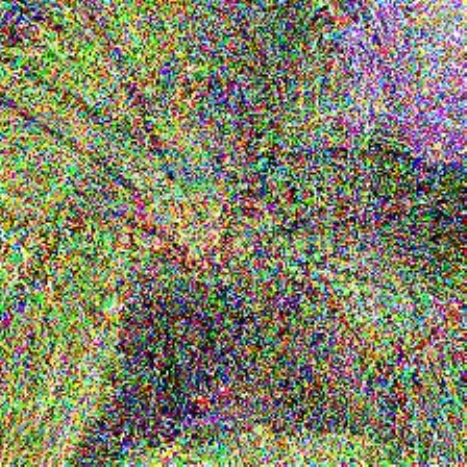}$$&
    \includegraphics[width=0.2\textwidth]{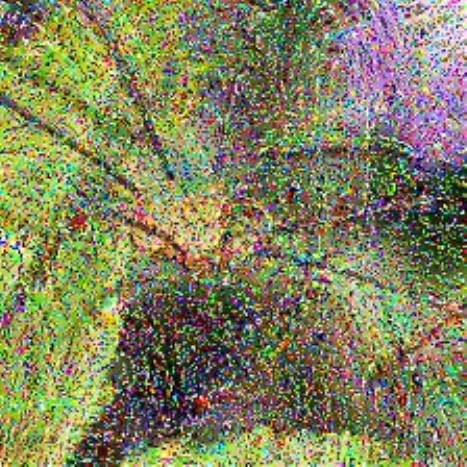}$$&
    \includegraphics[width=0.2\textwidth]{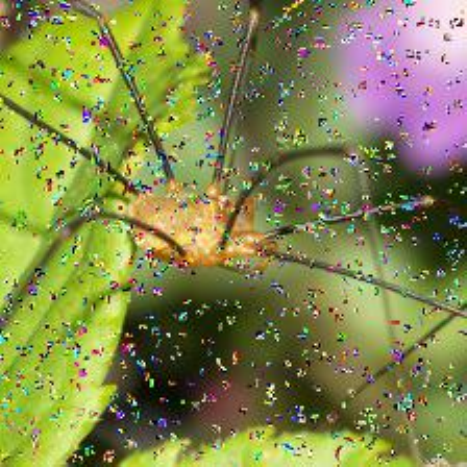}$$&
    \includegraphics[width=0.2\textwidth]{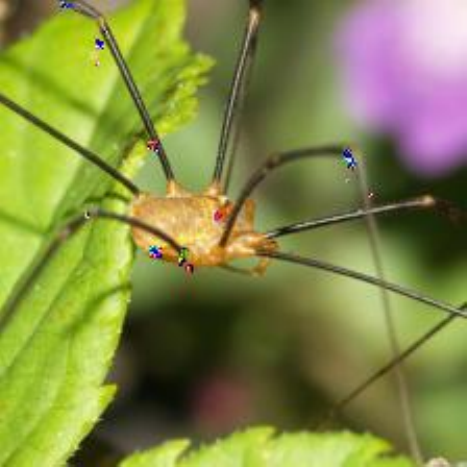}$$
    \\
    True label: \textbf{harvestmen}&
    Predicted label: \textbf{fountain}&
    Predicted label: \textbf{fountain}&
    Predicted label: \textbf{bubble}&
    Predicted label: \textbf{beetle}
    \end{tabular}
    }
    \caption{
    Flowchart of our sparse (top) and $2\times2$ patch (bottom) adversarial attacks trim process on Imagenet standard $Resnet50$ model, for attacks bounded to $\epsilon_0=224$. We present the adversarial inputs produced for distinct $\epsilon_0$ bounds during the process and the predicted label for each, compared to the true label.
    }
    \label{fig:trim_vis}
\end{figure*}




Let $\mathcal{X}\in[0,1]^n$ be some normalized data space comprising $N$ data points, and we denote $[N]\equiv\{i\}_{i=1}^N$. Let $x\in\mathcal{X}$ be a data sample and let $\delta\in\mathcal{X}$ be a perturbation, for $\delta$ to be applicable on $x$ it must be limited s.t. the perturbed data sample remains in the data space  $x_{\delta}=x+\delta\in \mathcal{X}$. Let $GT:\mathcal{X} \to \mathcal{Y}$ be a ground truth function over $\mathcal{X}$ and target space $\mathcal{Y}$, and let $M:\mathcal{X} \to \mathcal{Y}$ be a model aiming to predict $GT$. Given a data sample $(x, y)\in\mathcal{X}\cross\mathcal{Y}$, a criterion over the model prediction $\ell:\mathcal{Y}\cross\mathcal{Y}\to \mathcal{R}^+$, and $L_0$ norm bound $\epsilon_0\in[N]$, a sparse adversarial attack $A_s:\mathcal{X}\cross\mathcal{Y}\cross[N]\to\mathcal{X}$ targets the maximization of the criterion over the data sample and bound:
\begin{align}
    A_s(x,y,\epsilon_0) &= \arg \max_{\{\delta | x+\delta\in \mathcal{X}, \norm{\delta}_0 \leq \epsilon_0\}} \ell(M(x+\delta), y) \label{eq:sparse_adv_pert}
\end{align}



For a given choice of points and corresponding binary mask $B\in\{0,1\}^N$, the point-wise multiplication $\delta_s=B\odot\delta$ defines a projection onto the $L_0$ norm-bound space. We denote the set of binary masks with exactly $\epsilon_0$ ones as $C_{N,\epsilon_0}\subset\{0,1\}^N$ and, for $B\in C_{N,\epsilon_0}$, the $L_0$ norm of the resulting sparse perturbation $\delta_s$ is bound by $\norm{\delta_s}_0\leq \epsilon_0$. Sparse adversarial perturbations can be optimized using such projections \cite{fan2020sparse}. For an RGB normalized data space, we define the mask according to the pixels, i.e., $\mathcal{X}\in[0,1]^{H\times W\times 3}, N\equiv H\cdot W, C_{N,\epsilon_0}\subset\{0,1\}^{H\times W}$. Given an additional patch constraint with kernel $K\equiv (K_h, K_w)\in[H]\times[W]$, the perturbed points are limited to form exactly $\frac{\epsilon_0}{K_h\cdot K_w}$ patches of $K$'s shape, where we only consider accordingly divisible parameters. We denote the corresponding set of binary masks as $C^{K_h\times K_w}_{N,\epsilon_0}$. We allow for partial overlapping patches, as for sufficiently large kernels and $\epsilon_0$ bounds, most and then all of the binary masks $B\in C^{K_h\times K_w}_{N,\epsilon_0}$ will contain such. The patch adversarial attack is then denoted as $A_p:\mathcal{X}\cross\mathcal{Y}\cross[N]\cross[H]\cross[W]\to\mathcal{X}$ and we formulate the attacks targets as:
\begin{align}
    \delta_s\equiv A_s(x,y,\epsilon_0) =& \arg \max_{\{\delta_s=B\odot\delta| x+\delta\in \mathcal{X}, B\in C_{N,\epsilon_0}\}} \ell(M(x+\delta_s), y) \label{eq:binary_mask_adv_pert}\\
    \delta_p\equiv A_p(x,y,\epsilon_0,K_h,K_w) =& \arg \max_{\{\delta_s=B\odot\delta| x+\delta\in \mathcal{X}, B\in C^{K_h\times K_w}_{N,\epsilon_0}\}} \ell(M(x+\delta_s), y) \label{eq:patch_adv_pert}
\end{align}
\section{Method}
\label{sec:method}
\begin{figure}
 \centering
    \resizebox{\linewidth}{!}{
    \begin{tabular}{cc}  
    \includegraphics[width=\textwidth]{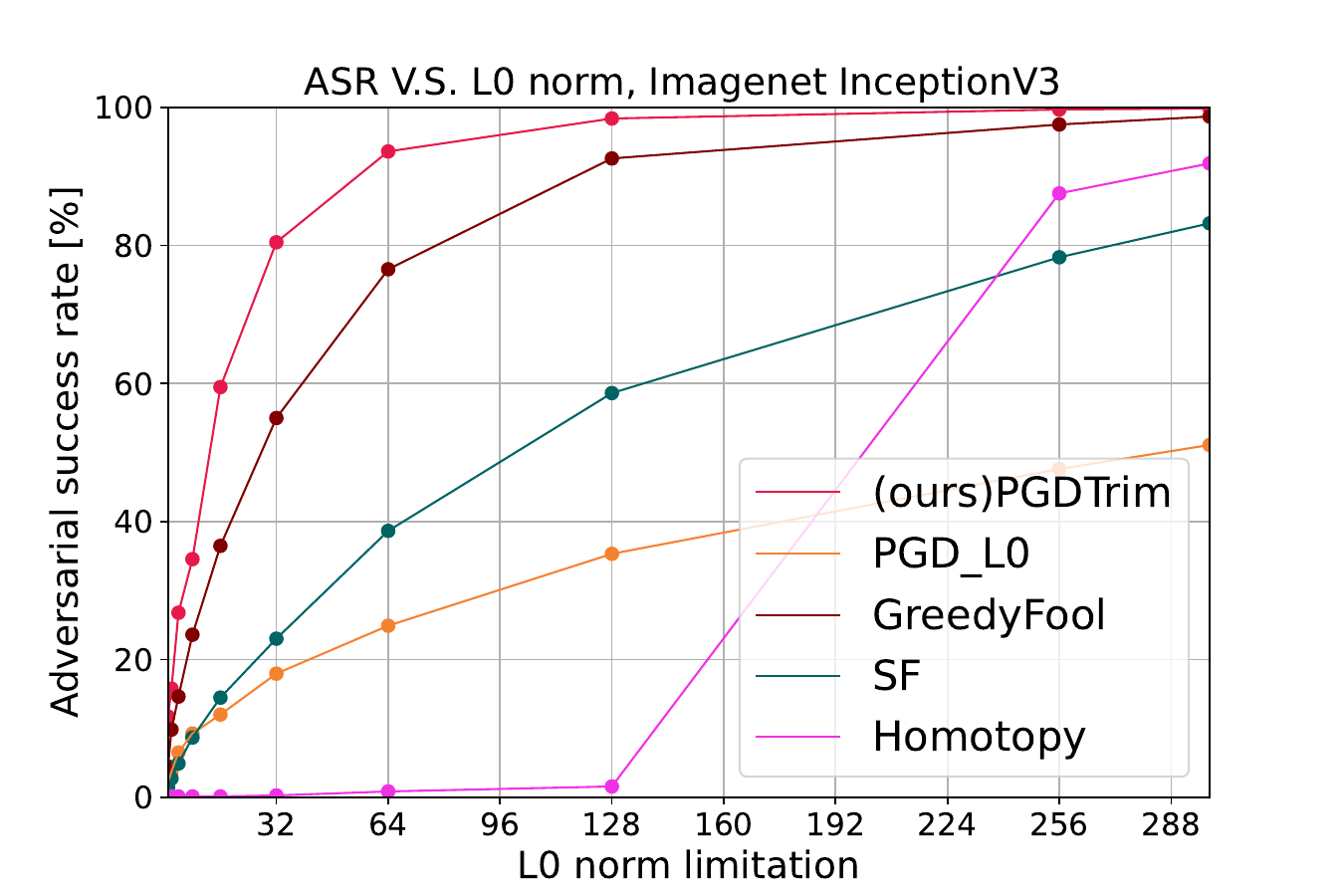}&
    \includegraphics[width=\textwidth]{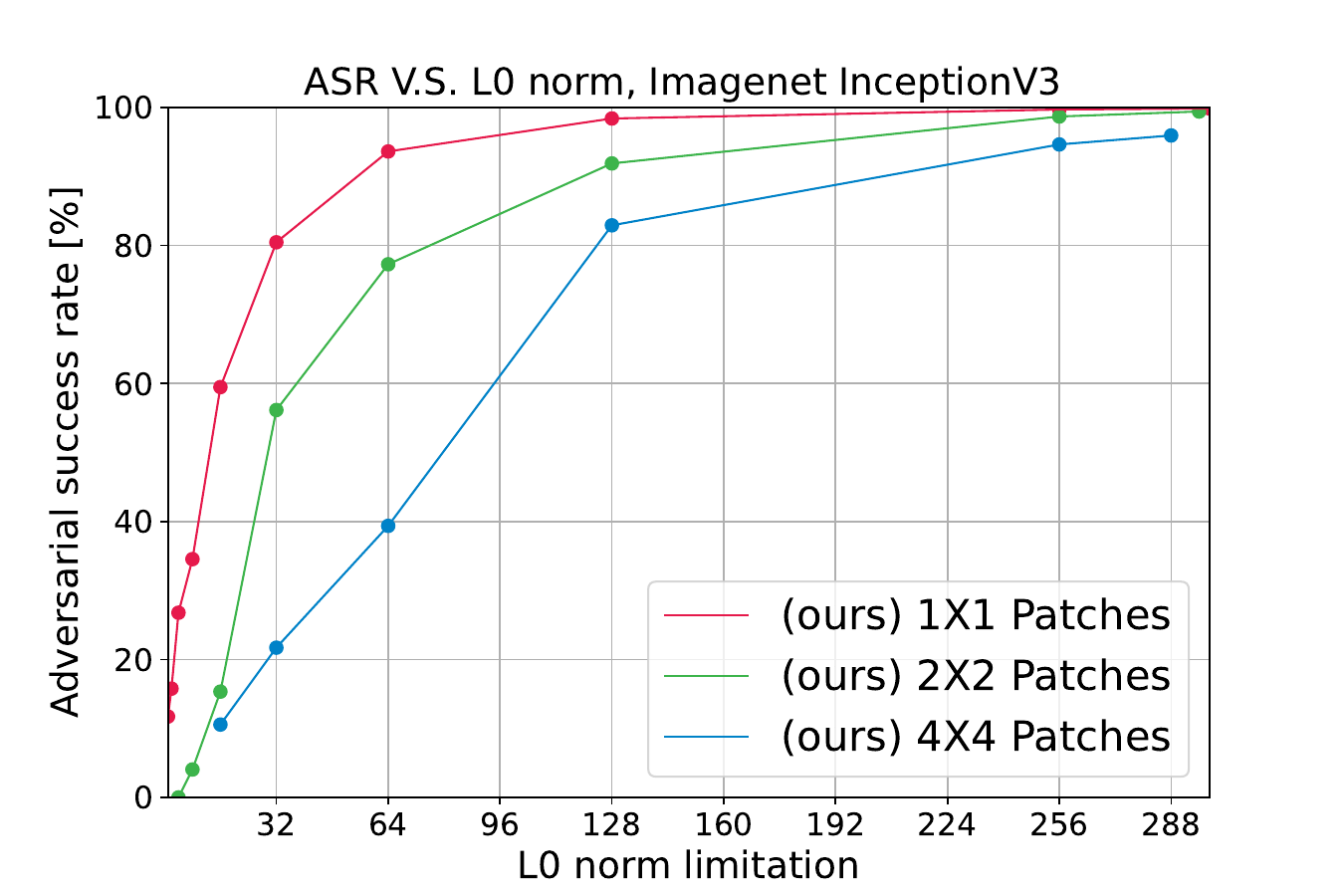}
    \end{tabular}
    }
 \caption{
 We compare our method to previous sparse attack works(left) and with various patch sizes (right) on the Imagenet dataset $InceptionV3$ model. We report the ASR as a function of $l_0$ for all attacks.
 }
\label{fig:imagenet_inceV3}
\end{figure}

\begin{figure}
 \centering
    \resizebox{\linewidth}{!}{
    \begin{tabular}{cc}  
    \includegraphics[width=\textwidth]{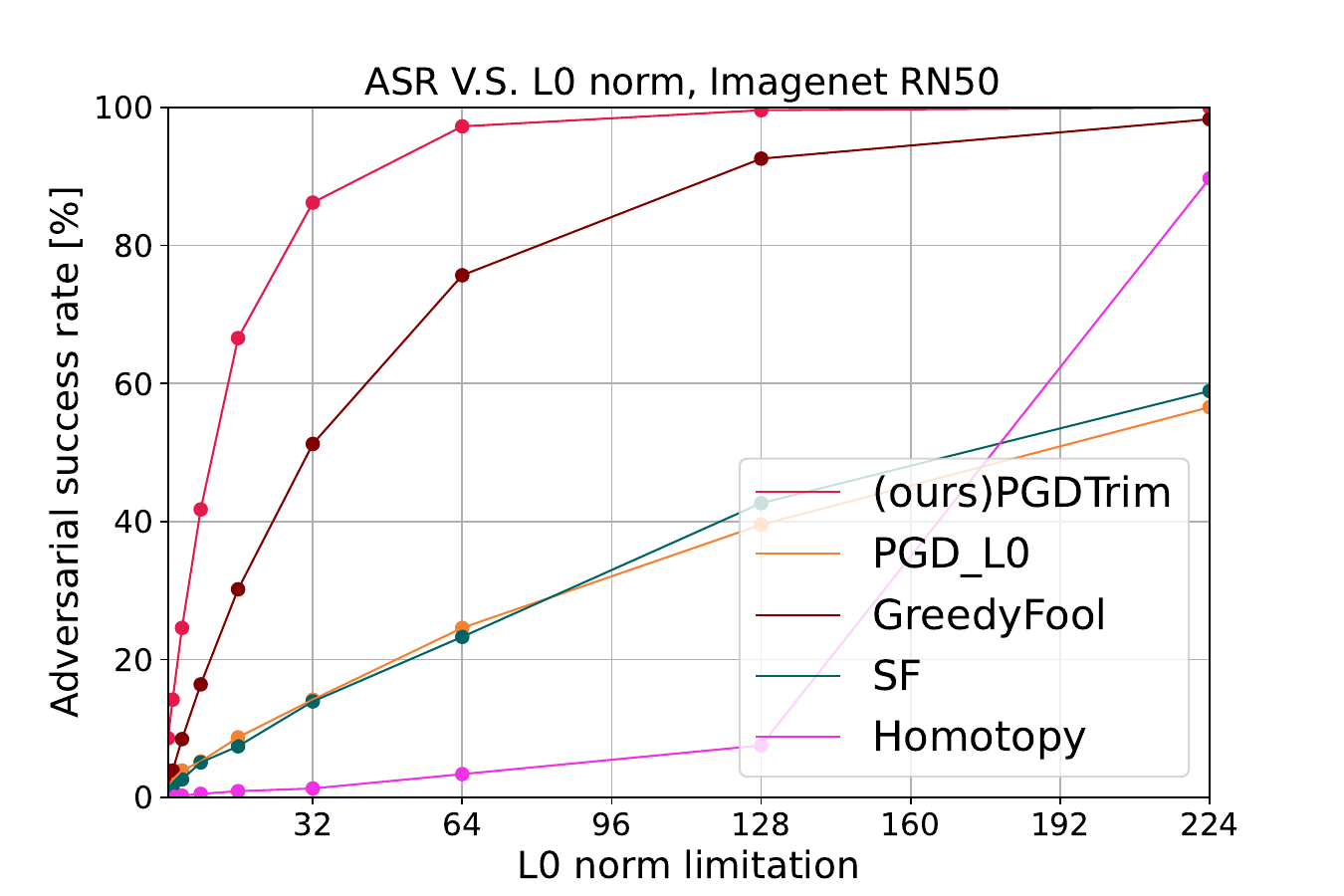}&
    \includegraphics[width=\textwidth]{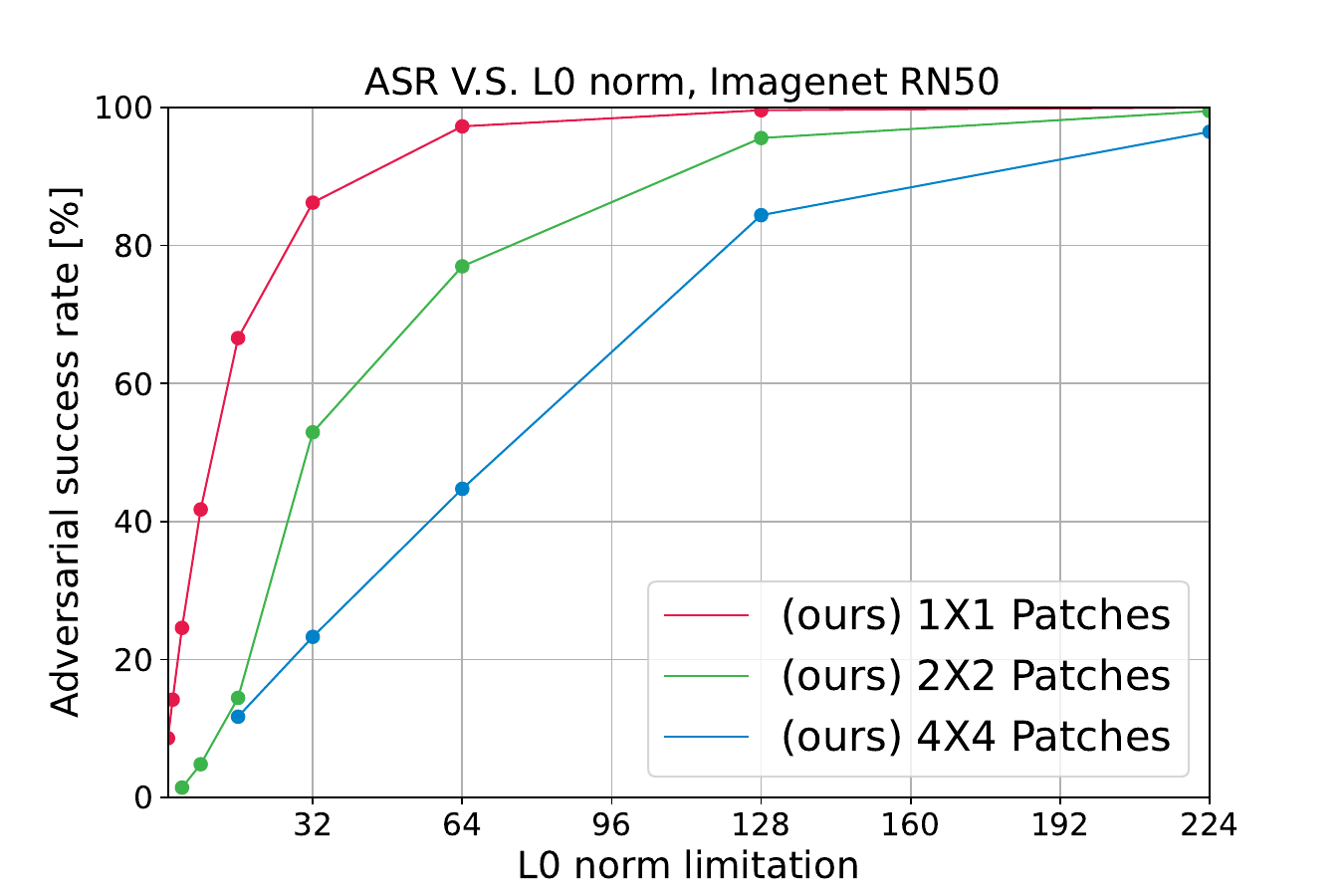}
    \end{tabular}
    }
 \caption{
 We compare our method to previous sparse attack works(left) and with various patch sizes (right) on the Imagenet dataset Resnet50 standard model. We report the ASR as a function of $l_0$ for all attacks.
 }
\label{fig:imagenet_SR50}
\end{figure}

\begin{figure}
 \centering
    \resizebox{\linewidth}{!}{
    \begin{tabular}{cc}  
    \includegraphics[width=\textwidth]{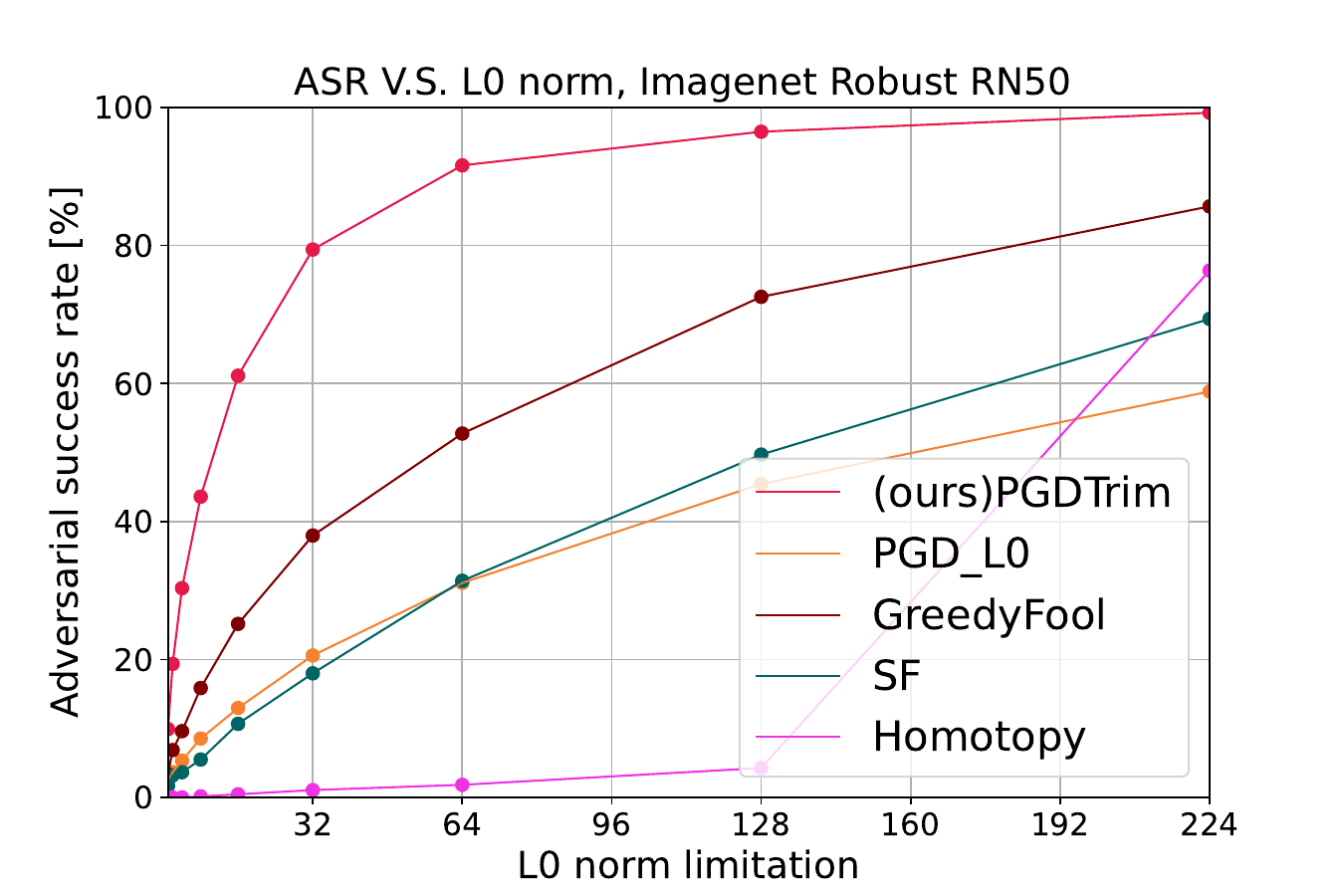}&
    \includegraphics[width=\textwidth]{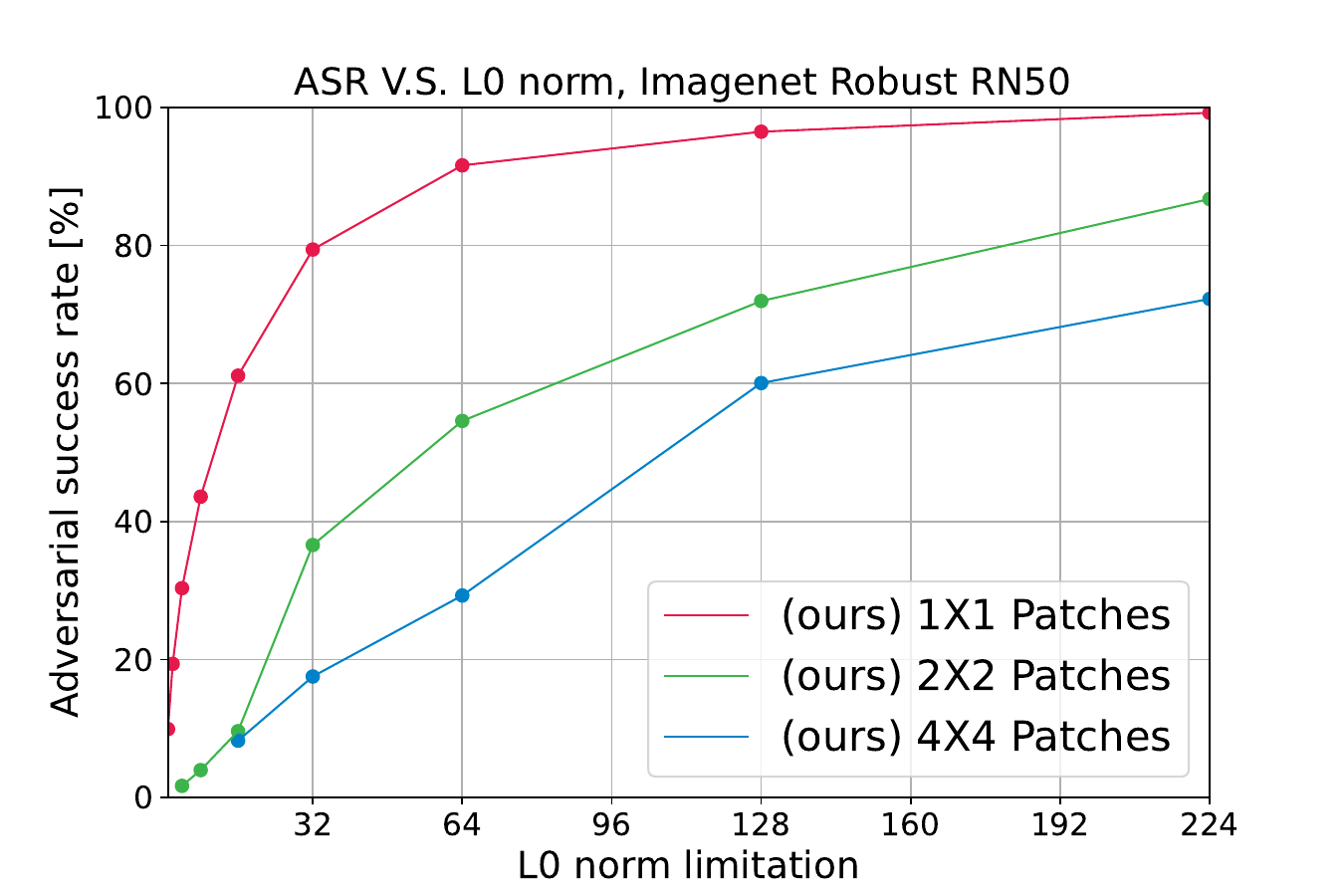}
    \end{tabular}
    }
 \caption{
 We compare our method to previous sparse attack works(left) and with various patch sizes (right) on the Imagenet dataset Resnet50 robust model. We report the ASR as a function of $l_0$ for all attacks.
 }
\label{fig:imagenet_RR50}
\end{figure}

\begin{figure}
    \centering
    \resizebox{\linewidth}{!}{
    \begin{tabular}{cc}  
    \includegraphics[width=\textwidth]{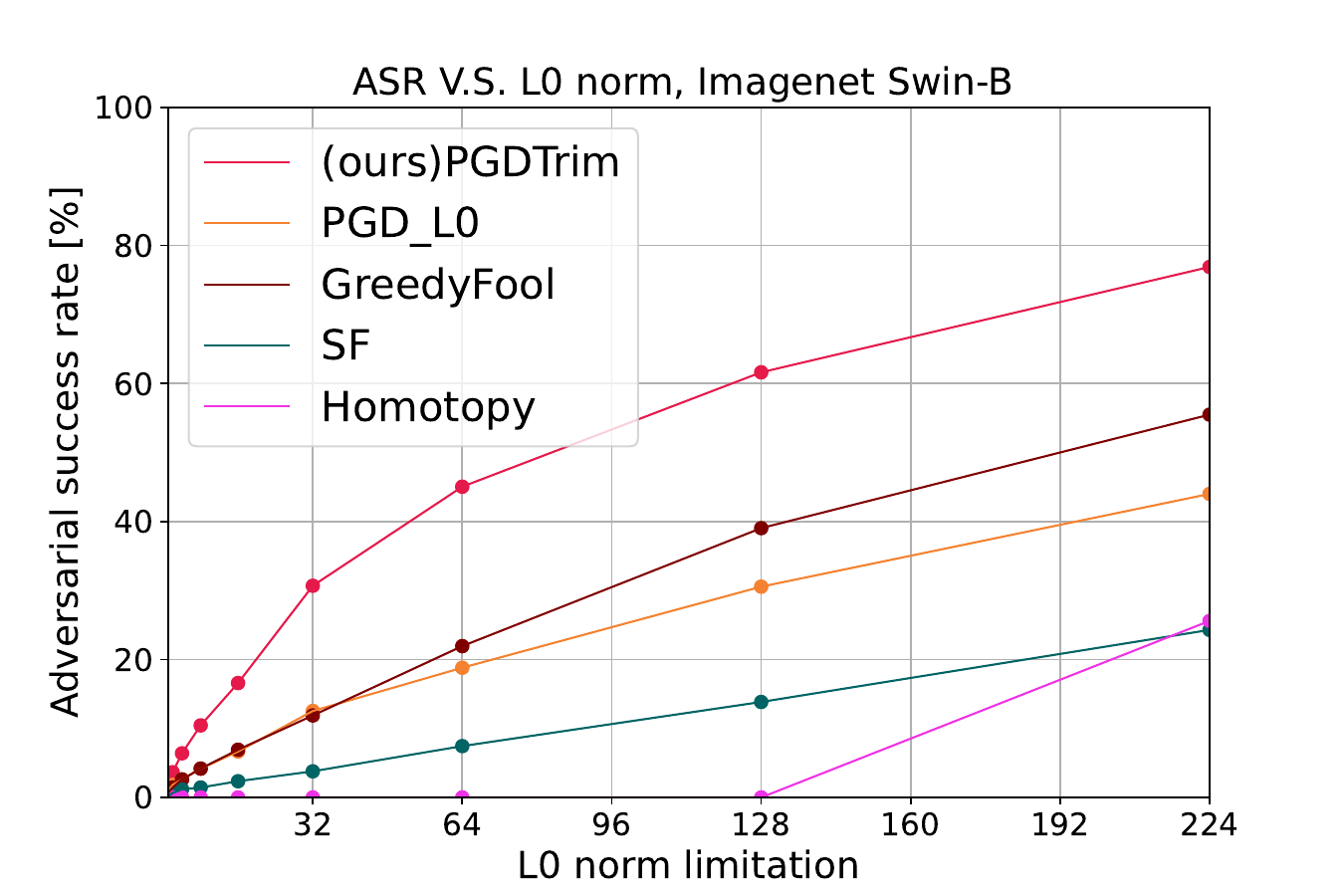}&
    \includegraphics[width=\textwidth]{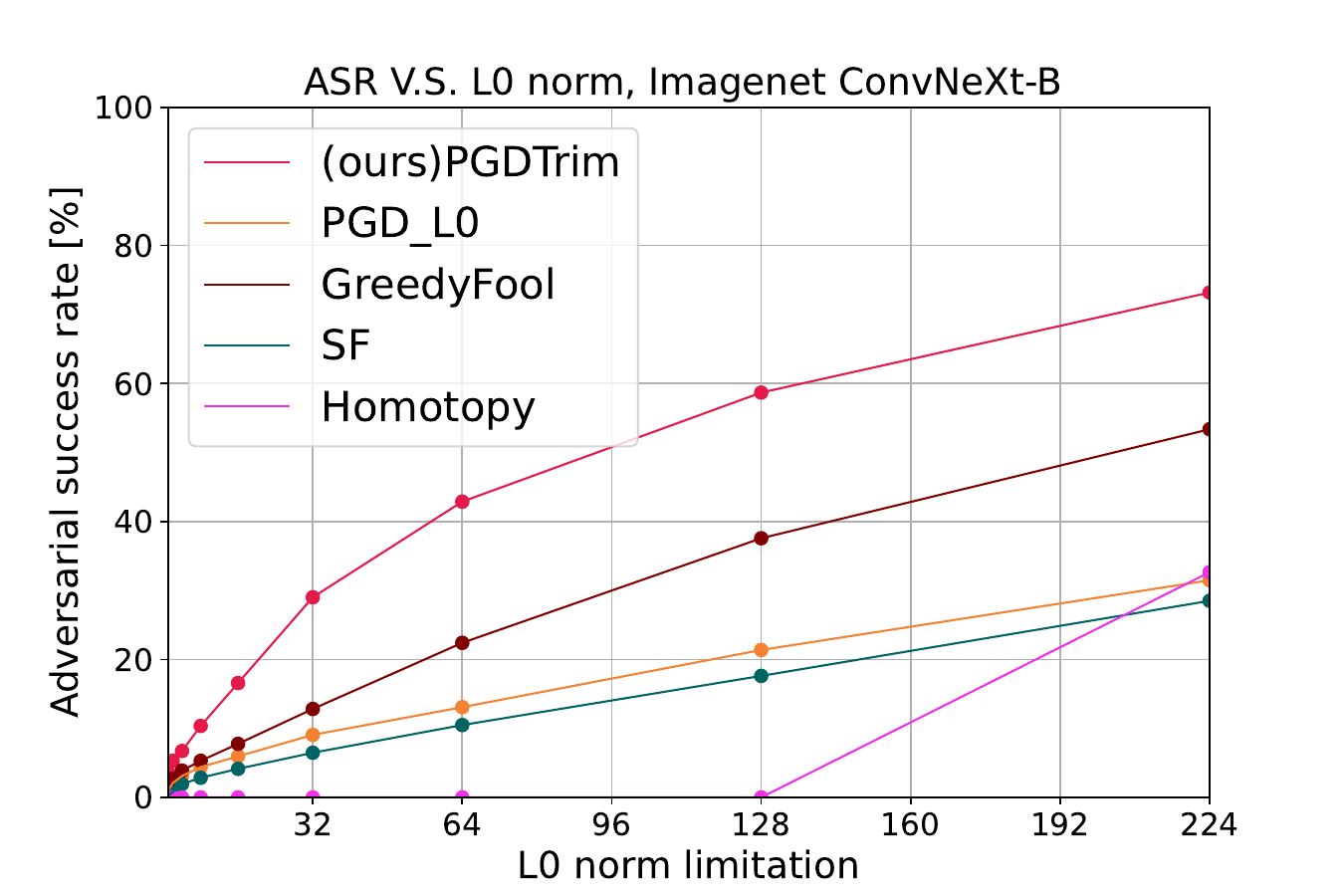}
    \end{tabular}
    }
    \caption{
 We compare our method to previous works on the Imagenet dataset, visual transformer-based SwinB model (left), and ConvNextB model (right). We report the ASR as a function of $l_0$ over sparse adversarial attacks.
    }
    \label{fig:imagenet_vit}
\end{figure}

We define our approach for point-wise evaluation and corresponding trimming of adversarial perturbations. We first present the process we denote as $TrimStep$ and discuss its optimization target and the point-wise evaluation criterion it utilizes. We discuss the underlying assumptions under which this process is most accurate. We then discuss our suggested sparse and patch adversarial attacks, which utilize the $TrimStep$ while aiming to fulfill the underlying assumptions.

\subsection{\itshape{TrimStep}}
Let $x, y, M, \ell, \epsilon_0$ be defined as in \cref{eq:sparse_adv_pert}, and let $\delta$ be a somewhat denser pre-optimized adversarial perturbation $\norm{\delta}_0 > \epsilon_0$. In this process, we aim to extrapolate a binary mask $B\in C_{N,\epsilon_0}$ from $\delta$ while targeting the proceeding optimization of a sparse perturbation under the fixed binary mask:

\begin{align}
    \delta_s^B &= \arg \max_{\{\delta_s=B\odot\delta| x+\delta\in \mathcal{X}\}} \ell(M(x+\delta_s), y) \label{eq:fixed_mask_adv_pert}\\
    B &= \arg \max_{B\in C_{N,\epsilon_0}} \delta_s^B \nonumber 
\end{align}

For this purpose, we consider the distributions of binary masks $B\in C_{N,\epsilon_0}$ and criteria $\ell(M(x+B\odot\delta), y),\ell(M(x+\delta_s^B)$ as prior and posterior distributions. We then define a point-wise evaluation criterion over the distributions and approximate the point-wise evaluation of the posterior by the prior. We denote the point-wise criterion over $\delta_s^B$  as $L_{\delta_s}\in\mathbb{R}^N$, and formally define the evaluation and its approximation:
\begin{align}
     L_{\delta_s} &= \mathbb{E}_{B\in C_{N,\epsilon_0}} \ell(M(x+\delta_s^B), y)\cdot B \label{eq:point_wise_crit}\\
     &\approx \mathbb{E}_{B\in C_{N,\epsilon_0}} \ell(M(x+B\odot\delta), y)\cdot B \label{eq:point_wise_crit_approx}
\end{align}

While computing $L_{\delta_s}$ directly is infeasible, we can efficiently compute the approximation given $\delta$. As the number of possible masks $|C_{N,\epsilon_0}|$ may be infeasible to compute, we further approximate this evaluation via Monte Carlo sampling. For each point in the data sample, the point-wise value of $L_{\delta_s}$ is the expectation of the attack target over binary masks that indicate the perturbation of the corresponding point. Accordingly, this evaluation estimates the expected benefit of each point selection to the attack target in \cref{eq:fixed_mask_adv_pert}. We, therefore, extrapolate the binary mask $B$ to perturb the top evaluated points, according to \cref{eq:point_wise_crit_approx}. Similarly, given an additional patch kernel constraint $K$ defined as in \cref{eq:patch_adv_pert}, the same process applies over the corresponding set of binary masks. Formally:
\begin{align}
    B_s &= \arg \max_{B\in C_{N,\epsilon_0}} L_{\delta_s}^T\cdot B \label{eq:fixed_mask}\\
    B_p &= \arg \max_{B\in C^{K_h\times K_w}_{N,\epsilon_0}} L_{\delta_s}^T\cdot B \label{eq:fixed_mask_patch}
\end{align}
The maximization in \cref{eq:fixed_mask} can be implemented directly as the top evaluated points in $L_{\delta_s}$; however, for \cref{eq:fixed_mask_patch}, we need to account for overlapping patches. We, therefore, use a max-out scheme when choosing the best patches, where the best patch in each step is chosen according to the sum of $L_{\delta_s}$ over the corresponding points. We then zero the $L_{\delta_s}$ values for the chosen patch to eliminate their benefit when considering overlapping patches. We can employ a similar process while applying a binary mask over the points in the kernel $K$ to allow for optimization of patches of any given shape. However, we consider this out of the scope of the current work. 

There are two approximations in the $TrimStep$ process. The first of which is approximating the best mask in \cref{eq:fixed_mask_adv_pert} as in \cref{eq:fixed_mask}, and the second is approximating the posterior in \cref{eq:point_wise_crit} via the prior in \cref{eq:point_wise_crit_approx}. We consider several assumptions for which these approximations should be most accurate. We first assume that attack criterion $\ell$ mainly depends on selecting significant points in the dense perturbation rather than a well-correlated group. Secondly, we assume that $\delta$ is sufficiently robust to the projections $B\odot\delta$, s.t., the decrease in the criterion for top evaluated points in $L_{\delta_s}$, $\ell(M(x+\delta), y)\to \ell(M(x+B\odot\delta), y)$ is mainly due to trimming less significant points. Finally, we assume that the $L_0$ gap between the perturbations $\Delta \epsilon_0 \equiv \norm{\delta}_0 - \epsilon_0$ is sufficiently small as it aids our previous assumptions. This entails that the point-wise significance should remain relatively unaltered between perturbations and limits the effect of the projections $B\odot\delta$. Under these assumptions, the top evaluated points according to $L_{\delta_s}$ should correlate well with the optimal mask selection in \cref{eq:fixed_mask_adv_pert}, and more so for sufficiently small $\Delta \epsilon_0$. Moreover, the top evaluated points in both \cref{eq:point_wise_crit_approx} and \cref{eq:point_wise_crit} should correlate to the points' importance in the $dense$ perturbation and, therefore, to each other. Thereby indicating the accuracy of the approximations in the $TrimStep$.

\subsection{\itshape{PGDTrim} and \itshape{PGDTrimKernel}}
\label{sec:attacks}
We continue to present our suggested sparse and patches adversarial attack based on the PGD iterative optimization scheme \cite{madry2018towards}. Both attacks use the same optimization scheme and differ only in utilizing the corresponding $TrimStep$. This optimization scheme aims to mitigate the inaccuracy of $TrimStep$ by fulfilling the underlying assumptions. The assumption on the attack criterion cannot be directly mitigated as it depends on the task; however, the other assumptions of small $\Delta \epsilon_0$ and robust $\delta$ are highly dependent on the optimization scheme. 
To fulfill the small $\Delta \epsilon_0$ assumption, we use a trimming schedule containing several applications of $TrimStep$ to gradually decrease the $L_0$ norm of the optimized perturbations until reaching the $\epsilon_0$ bound. We consider a logarithmic trimming schedule with up to $n_{trim}=\lceil log_2(N) \rceil - \lfloor log_2(\epsilon_0) \rfloor$ trim steps, where $N$ is the input size and $\epsilon_0$ is the $L_0$ norm bound. In addition, before each application of $TrimStep$, we optimize the current dense perturbation $\delta$ via the PGD scheme. To improve the robustness of $\delta$ to the perturbations $B\odot\delta$ we employ a corresponding dropout scheme. The dropout we consider in training the perturbations depends on the distributions of binary masks in the proceeding trim step. For a given current and following $L_0$ norms $L_0^{curr}, L_0^{next}$, the binary masks in the proceeding trim step are sampled from the set $B\in C_{L_0^{curr},L_0^{next}}$. We consider Bernoulli dropout from the corresponding distribution $Bernoulli(\nicefrac{L_0^{next}}{L_0^{curr}})$, as it best simulates the binary mask projection. We present a flowchart of our attacks in \cref{fig:trim_vis}, and in the supplementary material, we continue to discuss our optimization scheme and provide an entire algorithm of the resulting attacks.

\section{Experiments}
\label{sec:exp}



\paragraph{Experimental settings.}

We now present an empirical evaluation of the proposed method. We compare our method to previous sparse attacks on the $ImageNet$ classification task \cite{deng2009ImageNet} over various models. We present each attack's adversarial success rate (ASR), dependent on the $L_0$ norm bound, and show the result of our proposed method for both sparse and patch attacks. The $L_0$ norm bounds we consider are all values up to root input size $\epsilon_0=\sqrt{N}$, and we present the performance of the compared attacks for powers of $2$ in this range. The considered models are then the $InceptionV3$ \cite{szegedy2016rethinking}, standardly trained $Resnet50$ model \cite{koonce2021resnet}, adversarially robust $Resnet50$ model, and the visual transformer-based Swin-B \cite{liu2021swin} and ConvNeXt-B models \cite{liu2022convnet}. We use the pre-trained models made available by \cite{croce2020robustbench}, and the adversarially robust $Resnet50$ we consider is the corresponding state-of-the-art adversarial defense suggested by \citet{salman2020adversarially}, which we denote as robust $Resnet50$. The input size for the $InceptionV3$ model is then $N=299$, and $N=224$ for all other models.

In our method, for all the presented settings, we use $K=100$ PGD iterations for optimizing perturbations and $MC=1000$ Monte Carlo samples in our trim steps, where if these samples are sufficient, we compute the expression in \cref{eq:point_wise_crit_approx} directly. We compute the attacks for $n_{trim}=11$ trim steps and $n_{restarts}=11$ restarts; we use the PGD restarts optimization scheme to re-initiate the attack with fewer trim steps, as doing so will result in different perturbations and allow for re-evaluation of points trimmed in the extra steps. We use the default settings suggested by the authors for all the compared attacks for all the presented settings. In addition, as $GF$, $SF$, and $Homotopy$ attacks minimize the $L_0$ for each sparse adversarial perturbation instead of utilizing $\epsilon_0$ bounds, we report their ASR for each $L_0$ limitation as the rate of produced adversarial perturbations with correspondingly bounded $L_0$ norms.

\subsection{Experimental results}

In \cref{fig:trim_vis}, we show the trimming process of our sparse and patch attacks. We see that the perturbed points are gradually trimmed until reaching the $\epsilon_0$ bounds with the most significant points remaining.
In \cref{fig:imagenet_inceV3}, we compare the ASR of previous sparse attacks to our sparse and patch attacks on the $InceptionV3$ model. In this setting, our sparse attack achieves the best ASR on all the presented attacks and $100\%$ ASR starting from $\epsilon_0=128$. The second best sparse attack is $GF$, which shows comparable results to our patch attack over $2\times2$ patches. Our patch attacks over $4\times4$ achieve somewhat lower results, possibly due to the attacks' scope being more limited under this patch constraint.
In \cref{fig:imagenet_SR50}, we compare the ASR of previous sparse attacks to our sparse and patch attacks on the standard $Resnet50$ model. Similarly, our sparse attack achieves the best ASR and $100\%$ starting from $\epsilon_0=128$. Our results for $2\times2$ and $4\times4$ patches are again somewhat lower than our sparse attack, with the $2\times2$ setting comparable to the second-best sparse attack, $GF$.
In \cref{fig:imagenet_RR50}, we compare the ASR of previous sparse attacks to our sparse and patch attacks on the robust $Resnet50$ model. Our sparse attack again achieves the best ASR with $100\%$ achieved at $\epsilon_0=224$, corresponding to the model's robustness. Moreover, these results significantly outperform all other sparse attacks, which may entail that our method performs relatively better in robust settings. Our results for $2\times2$ and $4\times4$ patches are significantly lower than those of our sparse attack, yet the $2\times2$ setting is still comparable to the second-best sparse attack, $GF$.
In \cref{fig:imagenet_vit}, we compare the ASR of previous sparse attacks to our sparse attack on the $Swin-B$ and $ConvNeXt$ VIT models. Similarly, our sparse attack achieves the best ASR on all the compared settings and significantly outperforms other sparse attacks.

\section{Discussion}
\label{sec:conclusion}
This paper proposes novel sparse and patch adversarial attacks based on point-wise trimming of dense adversarial perturbations. For that purpose, we suggest ranking the points based on their average significance over potential resulting perturbations. We then approximate this significance based on the dense perturbation and choose the most significant points for our attacks under the corresponding constraints. Our sparse attack achieves state-of-the-art results for all the considered $L_0$ bounds. Moreover, our $2\times2$ patch attack shows results comparable to previous sparse attacks. The success of our method suggests that our point-wise evaluation may correspond to the significance of points in the input sample and not only in the adversarial perturbation. Therefore, our trimming-based approach is an efficient optimization method for sparse and patch attacks. In addition, our approach is the first to enable simultaneous optimization of multiple patches' locations and perturbations. Our approach does not require differentiability during trimming and applies to various real-world settings.



\bibliographystyle{unsrtnat}
\bibliography{egbib}

\appendix
\section{Adversarial attacks}
\label{sec:appendix_attack}
\subsection{Optimization scheme}
We continue to discuss the optimization scheme we use in the attack as described in \cref{sec:attacks}. We continue the discussion on the trimming schedule and offer continuous alternatives to the Bernoulli dropout. We have previously defined the number of trimming steps $n_{trim}$, and we now detail the logarithmic trimming schedule we consider. We first define the $L_0$ norm values to which we trim the perturbation in each step. The first perturbation we train is always whole $\norm{\delta_{init}}_0=N$, and the last is always constrained to $\epsilon_0$. For the maximal number of trim steps, the $L_0$ norms to which we trim and train perturbations are:
\begin{align}
    N, 2^{\lceil log_2(N) \rceil -1}, 2^{\lceil log_2(N) \rceil -2}, \ldots, 2^{\lfloor log_2(\epsilon_0) \rfloor + 1} ,\epsilon_0
\end{align}
For fewer trim steps, we skip a corresponding number of $L_0$ norms, where we attempt to keep the $L_0$ decrease ratio relatively fixed and otherwise slightly lower for the initial trim steps. In addition, we use the PGD restarts optimization scheme to re-initiate the attack with fewer trim steps, as doing so will result in different perturbations and allow for re-evaluation of points trimmed in the extra steps.

Concerning the continuous alternatives to the Bernoulli dropout, we consider the continuous Bernoulli and Gaussian dropouts, for which we preserve the mean as in the Bernoulli dropout and, when possible, the standard deviation.

\subsection{Attacks Algorithms}

We introduce algorithms for our sparse adversarial attack (\cref{alg:attack}), our patch adversarial attack (\cref{alg:patch_attack}), and the PGD-based optimization scheme they make use of (\cref{alg:pgd}). We first present the optimization scheme, which we denote as $Dropout-PGD(DPGD)$, then continue to present our sparse and patch attacks while using $DPGD$ as a procedure. Given a binary projection, dropout distribution, and initial perturbation, $DPGD$ optimizes a corresponding perturbation for maximized attack criterion. Our sparse and patch attacks then use $DPGD$ to optimize perturbations and then trim them using our point-wise evaluation. Given trim steps $L_0$ norms and dropout distribution class, our sparse attack utilizes $DPGD$ to optimize a corresponding perturbation in each trim step, then trim it to be the initial perturbation for the next step. 
Given an additional kernel constraint $K\equiv(K_h,K_w)$, our patch attack similarly optimizes and trims the perturbation but limits the resulting perturbation to consist of patches of $K$'s shape. Once the trimming process is finished, it returns the final binary mask, and an additional $DPGD$ procedure maximizes a corresponding perturbation. The $L_0$ bound is thereby specified in the norm of the last trim step.

\begin{algorithm}
	\caption{$Dropout-PGD(DPGD)$}
	\label{alg:pgd}
 	\hspace*{\algorithmicindent} \textbf{Input} $M$: attacked model\\
 	\hspace*{\algorithmicindent} \textbf{Input} $(x, y)$: input sample \\
	\hspace*{\algorithmicindent} \textbf{Input} $\ell$: attack criterion\\
	\hspace*{\algorithmicindent} \textbf{Input} $B$: Binary projection\\
 	\hspace*{\algorithmicindent} \textbf{Input} $\delta_\text{init}$: perturbation initialization \\
  	\hspace*{\algorithmicindent} \textbf{Input} $D$: dropout distribution \\
 	\hspace*{\algorithmicindent} \textbf{Input} $Iter$: number PGD iterations\\
	\hspace*{\algorithmicindent} \textbf{Input} $\alpha$: Step size for the attack
	\begin{algorithmic}[H]
        \State \underline{\textbf{initialize perturbation: }}
   		\State $\delta_\text{best} \gets \delta_\text{init}$
		\State $\text{Loss}_\text{best} \gets \ell(M(x+\delta_\text{best}), y)$
        \For{$k=1$ to $Iter$}
            \State \underline{\textbf{optimization step: }}
            \State $g \gets \nabla_{\delta} \ell(M(x+D(\delta)), y)$
            \State $\delta \gets \delta + \alpha \cdot B\odot\text{sign}(g)$
            \State $\delta \gets clip(\delta,-x,1-x)$
            \State \underline{\textbf{evaluate perturbation: }}
            \State $\text{Loss} \gets \ell(M(x+\delta), y)$
            \If{$\text{Loss} > \text{Loss}_\text{best}$}
                \State $\delta_\text{best} \gets \delta$
                \State $\text{Loss}_\text{best} \gets \text{Loss}$
            \EndIf
        \EndFor
        \State \textbf{return} $\delta_\text{best}$ 
    \end{algorithmic}
\end{algorithm}

\begin{algorithm}
	\caption{$PGDTrim$ sparse adversarial attack}
	\label{alg:attack}
 	\hspace*{\algorithmicindent} \textbf{Input} $M$: attacked model\\
	\hspace*{\algorithmicindent} \textbf{Input} $N$: input size\\
 	\hspace*{\algorithmicindent} \textbf{Input} $(x, y)$: input sample \\
	\hspace*{\algorithmicindent} \textbf{Input} $\ell$: attack criterion\\
	\hspace*{\algorithmicindent} \textbf{Input} $TrimSteps$: trim steps $l_0^{curr},l_0^{next}$ norms\\
 	\hspace*{\algorithmicindent} \textbf{Input} $Dropout$: dropout distribution class \\
   	\hspace*{\algorithmicindent} \textbf{Input} $MC$: number Monte Carlo samples\\
 	\hspace*{\algorithmicindent} \textbf{Input} $Iter$: number PGD iterations\\
	\hspace*{\algorithmicindent} \textbf{Input} $\alpha$: Step size for the attack
	\begin{algorithmic}[H]
        \State \underline{\textbf{initialize perturbation: }}
 		\State $B_\text{trim} \gets \{1\}^{N}$
  		\State $\delta_\text{best} \gets {\text{Uniform}(-1, 1)}^{N}$
		\State $\text{Loss}_\text{best} \gets \ell(M(x+\delta_\text{best}), y)$
		\For{$l_0^{curr},l_0^{next}$ in $TrimSteps$}
            \State \underline{\textbf{perturbation optimization: }}
    		\State $D \gets Dropout(l_0^{next}/l_0^{curr})$
            \State $\delta_\text{best}\gets\Call{DPGD}{M, (x, y), \ell, B_\text{trim}, \delta_\text{best}, D, Iter, \alpha}$
            \State \underline{\textbf{point-wise evaluation: }}
            \State $\text{BLoss} \gets \{0\}^{N}$
            \State $\text{BCount} \gets \{0\}^{N}$
            \For{$i=1$ to $MC$}
                  \State $B \gets {\text{Multinomial}(l_0^{next}, B_\text{trim})}$
                \State $\text{BLoss} \gets  \text{BLoss} + \ell(M(x+B\odot\delta_\text{best}), y)\cdot B$
                \State $\text{BCount} \gets  \text{BCount} + B$
            \EndFor
            \State $\text{BLoss} \gets  \text{BLoss} / \text{BCount}$
            \State \underline{\textbf{trim step: }}
            \State $B_\text{trim} \gets  \{0\}^{N} + B_\text{trim}[{\text{TopK}(l_0^{next}, \text{BLoss})}]$
            \State $\delta_\text{best} \gets B_\text{trim}\odot\delta_\text{best}$
            \State $\text{Loss}_\text{best} \gets \ell(M(x+\delta_\text{best}), y)$
        \EndFor
        \State \underline{\textbf{final perturbation optimization: }}
    	\State $D \gets \text{Identity}$
        \State $\delta_\text{best}\gets\Call{DPGD}{M, (x, y), \ell, B_\text{trim}, \delta_\text{best}, D, Iter, \alpha}$ 
		\State \textbf{return} $\delta_\text{best}$ 
    \end{algorithmic}
\end{algorithm}
 
\begin{algorithm}
	\caption{$PGDTrimKernel$ patch adversarial attack}
	\label{alg:patch_attack}
 	\hspace*{\algorithmicindent} \textbf{Input} $M$: attacked model\\
	\hspace*{\algorithmicindent} \textbf{Input} $N$: input size\\
 	\hspace*{\algorithmicindent} \textbf{Input} $(x, y)$: input sample \\
	\hspace*{\algorithmicindent} \textbf{Input} $\ell$: attack criterion\\
	\hspace*{\algorithmicindent} \textbf{Input} $TrimSteps$: trim steps $l_0^{curr},l_0^{next}$ norms\\
	\hspace*{\algorithmicindent} \textbf{Input} $K=(K_h,K_w)$: Kernel patch constraint\\
 	\hspace*{\algorithmicindent} \textbf{Input} $Dropout$: dropout distribution class \\
   	\hspace*{\algorithmicindent} \textbf{Input} $MC$: number Monte Carlo samples\\
 	\hspace*{\algorithmicindent} \textbf{Input} $Iter$: number PGD iterations\\
	\hspace*{\algorithmicindent} \textbf{Input} $\alpha$: Step size for the attack
	\begin{algorithmic}[H]
        \State \underline{\textbf{initialize perturbation: }}
 		\State $B_\text{trim} \gets \{1\}^{N}$
        \State $K_\text{size} \gets K_h\cdot K_w$
  		\State $\delta_\text{best} \gets {\text{Uniform}(-1, 1)}^{N}$
		\State $\text{Loss}_\text{best} \gets \ell(M(x+\delta_\text{best}), y)$
		\For{$l_0^{curr},l_0^{next}$ in $TrimSteps$}
            \State \underline{\textbf{perturbation optimization: }}
    		\State $D \gets Dropout(l_0^{next}/l_0^{curr})$
            \State $\delta_\text{best}\gets\Call{DPGD}{M, (x, y), \ell, B_\text{trim}, \delta_\text{best}, D, Iter, \alpha}$
            \State \underline{\textbf{point-wise evaluation: }}
            \State $\text{BLoss} \gets \{0\}^{N}$
            \State $\text{BCount} \gets \{0\}^{N}$
            \State $B_\text{kernel} \gets {\text{MaxPool}(B_\text{trim},K)}$
            \For{$i=1$ to $MC$}
                \State $B \gets {\text{Multinomial}(l_0^{next} / K_\text{size}, B_\text{kernel})}$
                \State $B \gets {\text{MaxPool}(\text{Pad}(B, ((K_h - 1,K_h - 1), (K_w - 1, K_w - 1))),K)}$
                \State $\text{BLoss} \gets  \text{BLoss} + \ell(M(x+B\odot\delta_\text{best}), y)\cdot B$
                \State $\text{BCount} \gets  \text{BCount} + B$
            \EndFor
            \State $\text{BLoss} \gets  \text{BLoss} / \text{BCount}$
            \State \underline{\textbf{trim step: }}
            \State $B_\text{trim} \gets  \{0\}^{N}$
            \For{$i=1$ to $l_0^{next}$}
                \State $B_\text{Max} \gets {\text{OneHot}(\text{ArgMax}(\text{SumPool}(\text{BLoss},K)))}$
                \State $B_\text{MaxKernel} \gets {\text{MaxPool}(\text{Pad}(B_\text{Max}, ((K_h - 1, 0), ( K_w - 1, 0)),K)}$
                \State $B_\text{trim} \gets  {B_\text{trim} + B_\text{MaxKernel}}$
                \State $\text{BLoss} \gets  {\text{BLoss}\odot (1 - B_\text{MaxKernel})}$
            \EndFor
        \State $\delta_\text{best} \gets B_\text{trim}\odot\delta_\text{best}$
        \State $\text{Loss}_\text{best} \gets \ell(M(x+\delta_\text{best}), y)$
        \EndFor
        \State \underline{\textbf{final perturbation optimization: }}
    	\State $D \gets \text{Identity}$
        \State $\delta_\text{best}\gets\Call{DPGD}{M, (x, y), \ell, B_\text{trim}, \delta_\text{best}, D, Iter, \alpha}$ 
		\State \textbf{return} $\delta_\text{best}$ 
    \end{algorithmic}
\end{algorithm}

\end{document}